# Data Mining to Measure and Improve the Success of Web Sites

Myra Spiliopoulou [*][¶]    Carsten Pohle [∥]

**Abstract**

For many companies, competitiveness in e-commerce requires a successful presence on the web. Web sites are used to establish the company's image, to promote and sell goods and to provide customer support. The success of a web site affects and reflects directly the success of the company in the electronic market. In this study, we propose a methodology to improve the "success" of web sites, based on the exploitation of navigation pattern discovery. In particular, we present a theory, in which success is modelled on the basis of the navigation behaviour of the site's users. We then exploit WUM, a navigation pattern discovery miner, to study how the success of a site is reflected in the users' behaviour. With WUM we measure the success of a site's components and obtain concrete indications of how the site should be improved. We report on our first experiments with an online catalog, the success of which we have studied. Our mining analysis has shown very promising results, on the basis of which the site is currently undergoing concrete improvements.

## 1 Introduction

The Web has the potential of an international marketplace. However, as pointed out by Chuck Williams in the San Francisco Examiner: "On the Internet, companies only have computers representing them. They better be intelligent computers." We are still far away from intelligent web sites in the conventional AI sense. But we do need to check continuously and improve the quality of a web site to the demands and expectations of its users – preferably on the fly.

We could study the satisfaction of visitors with a web site by selecting a representative user group and either study their behaviour or interview them directly, on the basis of pre-specified criteria that reflect the notion of satisfaction. This approach has certain drawbacks. First, the overhead of establishing an experimental environment is too high; regular success management cannot be afforded in such a way. Second, the goal of such an effort should be to maximize the success rather than merely compute a value for some success measure. Finally, selecting a representative user group is not trivial; the web is a global marketplace and cannot be trivially projected to a group of locally accessible users.

Hence, to improve the success of a site we need a different approach with the following properties: (i) It takes account of all visitors of the site; (ii) it is appropriate for being performed at a frequent, probably regular basis; (iii) it leads to concrete indicators of the site's shortcomings and of ways of alleviating them. For this purpose, we propose to improve success by analysing web usage patterns.

Data mining is by nature appropriate for this kind of analysis. The activities of all users are recorded in the web server log and the data mining paradigm provides the methodology for analyzing them. However, data mining is not per se adequate to improve success. We need an appropriate model of navigation behaviour, so that the discovered patterns can provide the indicators necessary for improving the site. We further need a measure of the site's success, which can be computed in the data mining process. We need a miner to perform navigation pattern discovery and a methodology for applying our measure to obtain the improvement indicators necessary.

---
[*]Institute of Information Systems,Humboldt University Berlin, Spandauer Str. 1, D-10178 Berlin
[¶]Contact author    Email: myra@wiwi.hu-berlin.de
[∥]Institute of Information Systems, Humboldt University Berlin, Spandauer Str. 1, D-10178 Berlin

In this study, we propose a complete model that measures and improves the success of a site by navigation pattern discovery. We first discuss related work in section 2. In section 3, we propose measures for estimating the success of a site in fullfilling the objective goals of its owner. In section 4, we present a model of navigation behaviour, on the basis of which success can be improved, and we shortly present the miner WUM [Spi99] that performs navigation pattern discovery. Section 5 describes the methodology of applying the miner to discover patterns for different user groups and of assessing the needed improvements through comparison among the patterns of the groups. In section 6, we apply our method to control the success of an online catalog. The last section contains a summary of our work and an agenda of further research issues.

## 2 Related Work

The interest in monitoring the usage of a web site is probably as old as the web itself. Early tools assisted web site administrators in studying and balancing the web server's load. Modern tools for web access monitoring support the computation of statistics that can serve as the basis for success analysis. Advances closely related to our work come from two domains: success measures for commercial web sites and data mining techniques for analyzing web usage.

### 2.1 Measuring the Success of a Site

The first efforts in modelling the success of a web site related to the quality of its pages. Sullivan distinguishes among (i) quality of service, such as response time, (ii) quality of navigation, expressed in the navigation modes supported by the site and (iii) accessibility of a page [Sul97]. However, these measures are difficult to quantify at the level of a whole site, especially because the importance of each page varies and is often context-sensitive.

In [Eig97], Eighmey presents an experimental setting for measuring the quality of commercial web sites. Quality is modelled as a set of factors like the information utility of the presented contents, ease of use and attractiveness of the presentation metaphor. These factors do not constitute measures but are appropriate for ranking. They were obtained from a sample of users that filled a questionnaire. Questionnaires are also used by Alpar et al. to evaluate the satisfaction of users with a number of web sites [Alp99]. In their study, user satisfaction is again measured on the basis of various factors that can be measured (like clicks performed or time spent in a site) or at least ranked. However, as noted in the introduction, such methods rely heavily on the selection of a representative sample of users and on the interaction with these users.

Drèze and Zufryden applied the method of conjoint measurement [GS78] to find a web site's most important attributes and their optimal shape from a user's point of view [DZ97]. They developed a prototype web site which presented different layouts to different users. From the web usage logs they computed the number of page requests and the duration of the visits, and used them as measures of the site's efficiency. This approach measures efficiency without involving the user. However, these measures do not reflect the success of the site towards the purposes it serves. A commercial site designed for the online ordering of products is not really successful if users are just browsing through its contents without ever purchasing anything.

The need of measuring the success of a site *with respect to the objective goals of its owner* is reflected in [BPW96]. Berthon et al. propose two measures of the site's success, the *contact efficiency* and the *conversion efficiency*. The first measure returns the fraction of users that spend at least a user-defined minimum amount of time exploring the site. The second measure returns the ratio of users that after exploring the site also purchased something. Hence, the success of the site is defined as its efficiency in "converting" visitors into customers and can be measured without the involvement of users.

The study of James Ho goes one further step in the direction of evaluating the success of a site on the basis of its goals [Ho97]. He proposes a framework for the evaluation of commercial web sites, in which three types of business purposes are distinguished: promotion of product and services, provision of data and information, processing of business transactions. Then, he introduces four factors of "value creation" from the visitor's point of view: timely, custom, logistic and sensational. However, these factors can only be ranked in interaction with a user.

All these studies focus on the mere measurement of a site's success. In contrast, the goal of our study is to measure *and improve* the site's success.

## 2.2 Advances in Web Usage Mining

Web usage mining encompasses studies in which knowledge is obtained through the analysis of web usage. This covers correlations among products or web pages, market segmentation on the basis of user demographics and interests, as well as analysis of a site's success.

In [ZEMK97, PE98], correlated but not linked web pages are discovered by clustering pages requested together by the site's visitors. This approach can be used to construct construct dynamic web pages automatically that provide links to pages considered relevant by earlier visitors [PE98].

Assistance to the novice user is similarly the goal in [Wex96, JFM97]. Wexelblat records the path followed by each user, identifies the most frequent paths among them and uses them to suggest an appropriate path through the site [Wex96]. A similar approach is used in [JFM97]. The rationale behind this approach is that if many users follow the same path in their search for information, this path should be suggested to unexperienced users, to help them in their search. While such recommendations may help new users in getting oriented, it does not take the objectives of the site itself into account, nor gives hints on how the site could be improved towards user groups who visit the site for different reasons and thus behave differently.

In [ZXH98, Mar99, BM98], OLAP technology is employed for prediction, classification and time-series analysis over web usage data. Zaiane et al. monitor the traffic in the web site and also analyse the evolution of user behaviour in terms of preferred pages, as the users grow more experienced. In the SurfAID project, a warehouse over web usage data is established and time series analysis is combined with association rules to discover unexpectedly evolving correlations among products [Mar99]. Büchner and Mulvenna propose the establishment of a warehouse, in which web usage data are combined with customer data, concept hierarchies on page contents and user demographics, as well as enterprise knowledge, e.g. in the form of previously discovered rules [BM98]. Although user activities form the basis of these types of analysis, the issue of improving the site itself is not addressed.

The discovery of web usage patterns with conventional mining techniques is proposed in [CPY96, CMS97, CTS99]. Chen et al. discover frequently accessed paths by applying a methodology similar to the discovery of association rules [CPY96]. Cooley et al. organize URL requests into user sessions [CMS99] and then apply association rule discovery and sequence mining to extract correlations among pages [CMS97, CTS99]. Wu et al. propose a similar approach for mining frequent traversal paths and groups of most frequently visited pages[WYB98]. In [PZOD99], Parthasarathy et al contribute an approach for mining dynamic databases more efficiently for sequences. However, in [Spi99] it has been shown that conventional mining algorithms are not appropriate for the discovery of web usage patterns, because (a) modelling navigation patterns as associations or sequences oversimplifies the problem and (b) statistical measures like frequency of access are too simple for navigation pattern discovery.

The miners MiDAS [BBA$^+$99] and WUM [SF99, Spi99] have been designed especially for the discovery of navigation patterns in the web. To alleviate the second shortcoming of conventional sequence miners, they are equipped with a mining language, in which sophisticated statistical and structural constraints can be experessed. However, for MiDAS a navigation pattern is still a sequence, while WUM models navigation patterns as arrays of *directed acyclic graph*s annotated with statistic data.

The different conception of navigation patterns between WUM and other sequence miners is due to the fact that they concentrate on patterns that reflect correlations among events (here: page accesses). WUM focuses rather on depicting and exploiting the navigation behaviour of user groups, in order to improve the web site accordingly. Our first results have shown that the model of navigation patterns is appropriate in this context [SPF99, SB00], but also that it must be accompanied by a model that measures and improves success and by a procedure for the mining process. In this study, we present the complete framework of modelling success and navigation behaviour and combining the two to improve the success of a site.

# 3 The Notion of "Success" for Web Sites

The measures of the quality of a site should be designed with respect to the business objectives of its owner. To model success in this context, we undertake three steps. We first model the contents of a site according to concepts reflecting its objective goals. We then categorize the site's users with respect to their activities in pursuing those goals. Finally, we define "site success" as the efficiency of its *components* in helping users to achieve the site's goals.

## 3.1 Objective Goals and Page Types Reflecting them

A web site may serve multiple purposes. A commercial web site for online merchandizing typically offers a search mechanism over its online product catalog and an ordering service for purchasing selected products. Additionally, a software vendor may also offer a chat forum, where clients can exchange experience and assist each other. A document archiver or a meta-crawler might adorn the search results with advertisement icons, thus serving two purposes: document search assistance and product marketing.

### 3.1.1 Specifying the Site's Goal

When measuring the success of a web site, the analyst must first specify the context, in which this analysis takes place, i.e. the site's goal towards which success should be measured. Clearly, a site for software merchandizing is very successful towards the goal of online purchases if people do buy software online, even if they never use the interaction forum.

The specification of the objectives towards which the analysis of the success factors should be performed corresponds to the problem specification step that precedes any further activity in the lifecycle of knowledge discovery. We assume that the analysis concerns *one objective at a time* and we characterize this objective as *"the goal of the site,"* as far as the analysis is concerned.

### 3.1.2 Pages Reflecting a Site's Goal

To make the site's goal explicit for the analysis of user behaviour, we characterize the site's pages in terms of their function in pursuing this goal.

**Definition 1:** An "action page" is a page whose invocation indicates that the user is pursuing the site's goal. A "target page" is a page whose invocation indicates that the user has achieved the site's goal.  □

In a merchandizing site, a filled query form for the catalog of products would be an action page, while the submitted product ordering form could be a target page. For a document archiver, an action page would be an invocation of the text retrieval service. The selection and inspection of a single document from the result list can be characterized as a target page.

We assume that a target page cannot be reached without accessing an action page first. This assumption is reasonable since obtaining a document or a product presupposes a mechanism for acquiring it from a collection.

In our definition of action and target pages, we observe a site from the viewpoint of the *services* it offers in the framework of its goal and propose a *service-based concept hierarchy* to model the different service options, parameters and combinations supported by the site.

### 3.1.3 Service-based Concept Hierarchies

Concept hierarchies are used in market-basket analysis to generalize individual products into more abstract concepts. This enables the discovery of correlations that are manifested frequently enough among the abstract concepts although they occur rarely among individual products. Büchner and Mulvenna also propose the establishment of concept hierarchies on the demographics of web users in order to study the segmentation of a company's electronic clientel [BM98].

For the measurement of success towards the site's goal we propose a different type of concept hierarchy. Instead of abstracting the site's URLs into concepts reflecting their contents, we suggest to model and abstract the site's *services* that generate the URLs and fill them with contents. For example, an online product catalog, a document archive or a database are not described in terms of the objects they contain but in terms of the search parameters and the permissible combinations they support for retrieving these objects. In this context, a page may contain different services, e.g., a homepage may contain a search service and an order button. Also, the same service may appear in mutliple pages, e.g., as a frame. In our evaluation, we observe the invocation of a service, as recorded in the log, and not the page contents.

An example service-based concept hierarchy is depicted in Fig. 1. It distinguishes action pages and target pages. The former are query services for the retrieval of lists of objects, while the latter are descriptions of individual objects; in another example, target pages would be order forms for individual objects. For action pages, we model the search strategies as combinations of individual search parameters, which can be typed or invoked as buttons. The results of a search can vary in length and in form of presentation. Target pages are modelled similarly, by distinguishing among different formats of displaying a single object, e.g. concise and detailed description.

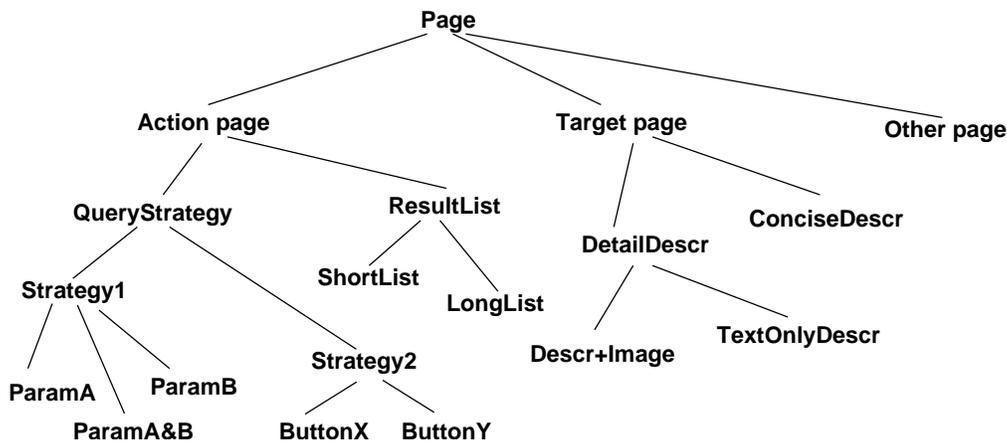

Figure 1: Example of a service-based concept hierarchy

By mapping URLs into abstract service-based concepts, we can map the site's pages into action and target pages at different levels of abstraction and observe the behaviour of its users in accessing these pages. Thus, the obvious fact that rarely do two distinct users request the same document or the same product from a large catalog is alleviated in a way conformant to our task of studying the site's usage rather than the correlation among products offered by it.

The additional overhead of establishing service-based concept hierarchies is payed off by the possibility to analyse navigation patterns that are insignificantly rare at the level of individual pages. This overhead is much lower than the overhead of building product hierarchies in market-basket analysis because the parameters describing the services of a site are explicit and have less semantic dimensions than are conceivable for the products offered by a site.

### 3.2 Success as Contact and Conversion Efficiency

In the framework of web-based marketing, Berthon et al. anticipate that the success of a site is measured by the percentage of its visitors that get engaged in exploring it ("contact efficiency") and in the percentage of the visitors that finally become customers ("conversion efficiency") [BPW96]. They consequently categorize the site's visitors on the basis of whether they have performed activities conformant to the site's goal – in their case, the purchase of products. Thus, a "short-time visitor" is a user that reaches but soon abandons the site without exploring it, while an "active investigator" stays longer and explores the site. A subset of active investigators become "customers", i.e. they order products or perform some similar activity.

We first switch from the concept of *user* to the concept of *session*, since a user may launch multiple sessions pursuing different goals. We then generalize the aforementioned user types into session types that reflect visitor behaviour towards arbitrary site goals. Finally, we introduce respective efficiency measures, not for the site as a whole, but for its components.

### 3.2.1 Pursuing the Site's Goal Within a Session

We characterize all users accessing the site as "visitors." A sequence of activities performed by the visitor and observed by the analyst as a single work unit is termed a "session." The problem of establishing sessions is discussed in section 4.

**Definition 2:** An "active session" is a session containing at least one activity towards fullfilling the site's goal. All other sessions are termed "inactive." □

According to the definition of *action pages* in Def. 1, active sessions are those containing an access to at least one action page.

Our definition has two advantages in comparison to the distinction between active investigators and short-time visitors of Berthon et al [BPW96]. First, we can uniquely determine whether a user's session is active or inactive, without relying on criteria like length of stay or number of pages requested. Such criteria are only modestly reliable, because they might lead to misclassifying an experienced customer as a short-time visitor or a disoriented user as an active investigator. Second, the distinction between active and inactive sessions is done on the basis of the site's goal. The analysis of the site towards another goal only implies specifying the corresponding action and target pages: the active and inactive sessions are then automatically re-defined.

**Definition 3:** A "customer session" is a session in which the user has achieved the site's goal. □

According to Def. 2 and to our assumption that a target page is only reachable via some action page (see 3.1.2), a customer session is always an active session. We term all active sessions that are not customer sessions as "non-customer sessions." Similarly to active and inactive sessions, a session can be uniquely characterized as a customer session from the pages it contains.

In the following, we characterize the users that launched these sessions as "customers" and "non-customers" respectively. We do this for simplicity of formulation, notwithstanding the fact that the same physical person may behave once as a customer and once as a non-customer.

### 3.2.2 Contact Efficiency Inside a Site

Using the concepts of action page and active session as a basis, we define the "contact efficiency of an action page" as the ratio of sessions containing this page to all sessions in the log. It should be stressed that the log is not a set but a multiset, since it is possible that multiple users have performed the same sequences of activities, thus corresponding to identical sessions.

**Definition 4:** Let $Sessions$ denote all sessions recorded in the log and let $A$ be an action page of the web site. Then, the contact efficiency of $A$ is given by:

$$contacteff(A) = \frac{card(\{\{s \in Sessions | A \in s\}\})}{card(Sessions)} \quad (1)$$

where $card(\cdot)$ denotes cardinality and $\{\{\ldots\}\}$ denotes a multiset. □

The sessions counted in the numerator of $contacteff(A)$ are obviously active sessions, since $A$ is an action page. Thus, the contact efficiency of $A$ is the percentage of sessions in which an attempt to reach the site's goal was made *using action page $A$*. By computing this value for each action page, we can (i) identify the impact of each page on the overall success of a site in engaging visitors and (ii) detect pages with low contact efficiency.

**Definition 5:** The "relative contact efficiency of an action page" $A$ is the ratio of sessions containing this page to the cardinality of the multiset of active sessions, called $aSessions$:

$$Rcontacteff(A) = \frac{card(\{\{s \in aSessions | A \in s\}\})}{card(aSessions)} \quad (2)$$

In this definition, the numerator is the same as in Eq. 1, since a session containing $A$ is indeed an active session. This measure expresses the relative importance of each action page within a site and is appropriate for sites with many action pages and/or a large number of inactive sessions.

#### 3.2.3 Conversion Efficiency Inside a Site

Similarly to the contact efficiency of an action page, we define the conversion efficiency of a(n arbitrary) page towards a target page. In this definition, we also need to consider the paths used to reach the target page. For instance, it is important whether the target page was reached in 3 or 13 steps. If the site is designed as a flat hierarchy of pages, in which each important object should be reachable within a small number of steps, long paths to a target page are undesirable. On the other hand, if the objective goal of the analysis is the exposure of the user to advertisments, long paths may be more desirable than short ones.

**Definition 6:** We define the "conversion efficiency of a page $P$ towards a target page $T$ over a group of paths $\mathcal{G}$ from $P$ to $T$" as the ratio of the cardinality of $\mathcal{G}$ to the cardinality of all active sessions containing $P$:

$$conveff(P, T, \mathcal{G}) = \frac{card(\mathcal{G})}{card(\{\{s \in aSessions | P \in s\}\})} \qquad (3)$$

Here, a path is part of a session, composed of consecutive accesses. □

The paths $\mathcal{G}$ are parts of active sessions, since they contain $T$, a target page. Since they also contain $P$, the numerator is at most equal to the denominator.

This measure estimates the success of an arbitrary page in helping/guiding the users towards a target page. Our generic definition allows the estimation of different conversion efficiency values, e.g. over long and over short paths or over all paths, a value we denote as $conveff(P, T, *)$. With this measure, we can study the impact of each page in the success of the site and identify pages that have a low conversion efficiency and require improvement. However, to do so, we must identify the groups of paths over which the conversion efficiency should be computed: these groups are navigation patterns reflecting the users' behaviour and should be discovered by a miner. Navigation pattern discovery is discussed in section 4.

**Example 1:** We would like to compute the conversion efficiency of a page $P$ towards a target page $T$ over all paths. Page $P$ appears in 100 active sessions. In 20 sessions among them, page $A$ was accessed after $P$, then $T$ was requested. In 30 other sessions, page $B$ was accessed after $P$; in only 10 of these 30 sessions was $T$ accessed afterwards. In the remaining 40 sessions, the page accessed after $P$ was $C$; these users never reached $T$.

From the paths $PAT, PBT, PC$, only $PAT$ and $PBT$ involve both $P$ and the target page $T$. Thus, the value of the numerator is the number of times $PAT$ and $PBT$ were completely traversed. $PAT$ was traversed in 20 sessions, $PBT$ in 10 only. The denominator's value is 100, the number of active sessions containing $P$. Thus, $conveff(P, T, *) = \frac{20+10}{100} = 0.3$.

This value is rather low, at least in the context of some applications. To identify whether improvements are needed at page $P$ or at another page accessed after $P$, we must inspect all paths emanating from $P$ and leading to $T$ and identify the pages at which users gave up or followed other routes. This is part of the postmining phase, as described in section 5. △

### 3.3 The Knowledge Discovery Process for Success Analysis

The knowledge discovery process is typically modelled as a series of steps, namely (i) specification of the problem, (ii) gathering and preparation of the relevant data, (iii) analysis of the data with data mining techniques, (iv) evaluation of the results according to previously established measures, (v) interpretation of the results and (vi) action according to strategic decisions. One more step, the verification of the impact of the undertaken actions completes the process.

For the particular problem of improving the success of a web site, we model the aforementioned steps as follows:

1. Formal problem specification
   The goal of improving the site's success is modelled by the concepts of contact efficiency for action pages and of conversion efficiency towards target pages.

2. Data preparation
   For the concrete web site, the notions of "active investigator" and "customer" must be determined. Also, the pages must be selected that serve as "action pages" for this web site or "target pages" respectively.

   The individual pages can be abstracted into more generic concepts by establishing service-based concept hierarchies that model the services provided by the site. In that case, the page requests in the web server log should be replaced by requests to the abstract concepts.

   Finally, the web server log must be cleaned and the entries must be grouped to form visitor sessions. Data mining is applied on these sessions.

3. Data mining
   Web usage mining for the analysis of a site's success translates into the discovery of navigation patterns that reflect the contact, resp. conversion efficiency of frequently accessed pages.

   In the next section, we present the base principles of navigation pattern modelling and the mechanism of pattern discovery with the dedicated miner WUM.

   Navigation pattern discovery is performed on the portion of the web server log that contains the customer sessions. The discovered patterns reflect the *desired* behaviour of the visitors. These patterns are then used as a basis to analyze the sessions in the rest of the log, comprising the sessions of the active investigators that did not become customers.

4. Evaluation and interpretation of the results
   The measures for the evaluation of the discovered patterns are the contact and conversion efficiency introduced in this section. The interpretation of the results is based on the study of the patterns' contents and of their statistics and on the background knowledge of the site's owner.

5. Acting upon the mining results
   In the context of improving success, data mining should lead to concrete suggestions for the re-design of (part of) the site. Once this has been performed, the impact of the changes should be verified by analyzing the web server log obtained after the re-design.

In the next section, we discuss the theoretical aspects of the navigation pattern discovery process. Section 6 shows how the aforementioned steps of data mining, result evaluation, interpretation and action are applied to improve the success of a web site.

# 4 Navigation Pattern Discovery

The efficiency measures introduced in the previous section estimate the efficiency of the individual pages, which is indirectly reflected in the behaviour of the site's visitors. This behaviour is registered by the web site server in the form of consecutive URL requests. This log of individual requests must be transformed into a log of sessions, from which navigation patterns should be extracted. Efficiency estimation is then performed on the basis of these patterns.

The transformation of the web server log into a log of sessions appropriate for mining and the process of navigation pattern discovery are performed in the framework of the Web Utilization Miner WUM [1]. We provide here a short overview of the transformation and the mining phase of WUM, from the viewpoint of modelling and extracting patterns. A complete presentation of the theory and the architecture of WUM can be found in [Spi99, SFW99] and [SF99] respectively.

---
[1] `wum.wiwi.hu-berlin.de`

## 4.1 A Model of Web Usage Patterns

The basis data from which web usage patterns are extracted are the individual URL requests performed by each site visitor.

### 4.1.1 Sessions

A "session" is a sequence of consecutive URL requests performed by the same visitor. We assume that a mechanism distinguishing among sessions of different visitors is available. Many sites distinguish among visitors by requiring authentication or using cookies, while others use dedicated application servers that assign session identifiers to visitors automatically. For sites not using such mechanisms, a suite of heuristics proposed by Cooley et al [CMS99] can be applied.

The boundaries of a session can be specified either by duration or by content. We use the former option as supported by WUM [SF99]: the boundaries of a session are defined by placing an upper limit on either its total duration or on the duration of a stay on a page. Page access statistics may be exploited to obtain appropriate estimates for the upper limit value. Alternative specifications based on content, as proposed in [CY96, CMS99], have the disadvantage that they make assumptions on how the users navigate in the site and what they are supposed to access. This is feasible at most for customer sessions in our approach.

The establishment of sessions is coupled with the exploitation of concept hierarchies abstracting the individual URLs of the site. As proposed in 3.1.3, dynamically generated URLs can be properly modelled with a service-based hierarchy that describes the services that generate the pages. We thus replace individual URLs with abstract service descriptors that reflect the search strategies, format types and layouts used by the visitors. In the following, we use the generic term "page" to refer to an access request, which can be a URL or a more abstract concept.

We distinguish among different invocations of the same page within a session, by modelling the session as *a sequence of (page, occurence number) pairs*. We call a session element a "page occurence."

**Example 2:** We consider a fictitious online catalog of products, the URLs of which are abstracted to the concepts in the leaf nodes of the service-based hierarchy in Fig. 1. We assume the following sessions, modelled as sequences:

| 1 | (ParamA,1)(ShortList,1)(ShortList,2)(TextOnlyDescr,1)(TextOnlyDescr,2) |
|---|---|
| 2 | (ParamA,1)(LongList,1)(ParamA&B,1)(LongList,2)(TextOnlyDescr,1) |
| 3 | (ParamA,1)(LongList,1)(ButtonX,1)(LongList,2) |

All three sessions are active sessions since they contain accesses to pages categorized as action pages in the concept hierarchy of Fig. 1.

In session 1, the user issued a query by specifying a value for parameter `ParamA`. She obtained the results in the `ShortList` format and reached the target page `TextOnlyDescr` after browsing through two pages of results. This user actually found two objects of interest, reflected by two accesses on the target page.

The user of session 2 also issued a query using `ParamA`. However, after obtaining the first page of results in `LongList` format, she refined the search by specifying both parameters A and B (`ParamA&B`). She again obtained a page of results in `LongList` format, from which she reached the target page `TextOnlyDescr`.

The user of session 3 did not reach a target page. After issuing a query using `ParamA` and obtaining a list of results in `LongList` format, she switched to another search strategy, invoked by pressing `ButtonX`. After inspecting the first page of the corresponding list of results, the user abandoned the site. △

### 4.1.2 Generalized Sequences and Navigation Patterns

A session describes the activities of one user. In our analysis, we are interested in frequent or otherwise interesting behavioural patterns that represent multiple users. In conventional sequence mining, navigation patterns are modelled as sequences of events that occur in order but not necessarily consecutively [AS95]. In [Spi99, SPF99], we argue that this representation does not suffice to model the navigational behaviour of users. In particular, we are not interested only in sequences of pages frequently accessed in that order but also in identifying and inspecting the frequent and less frequent paths used to reach them. We thus proposed the notion of generalized sequence instead [Spi99]: A generalized sequence or "g-sequence" is a vector comprising page occurence and wildcards. The "navigation pattern of a g-sequence" is then the group of subsequences matching the g-sequence, internally represented as an array of trees.

**Example 3:** We use the three sessions presented in Example 2. We are interested in the navigation pattern of the users who started by specifying `ParamA` and reached the target page `TextOnlyDescr` after at most 3 steps. In WUM, this g-sequence $g$ is expressed as (`ParamA,1`) $[0;3]$ (`TextOnlyDescr,1`), where $[0;3]$ denotes a wildcard to be matched by zero to three page occurences [SFW99].

We can see that $g$ is matched by the first and second sequence in Example 2. Thus, these two sequences constitute the navigation pattern of $g$. By contrast, the g-sequence (`ParamA,1`) $[0;2]$ (`TextOnlyDescr,1`) is matched by the first sequence only. △

In pattern discovery, we are not interested solely in the contents of the patterns, but also in their statistics: The importance of a pattern depends also on whether it is frequent or rare.

**Definition 7:** Let $U$ be the set of (page, occurence number) pairs recorded in a web site. Let $\mathcal{L}$ be a sequence log over elements of $U$ and let $g$ be a g-sequence over elements of $U$ as well . The "hits" of $g$, $hits(g)$, is the number of sequences in $\mathcal{L}$ that are matched by $g$. □

**Definition 8:** Let $\mathcal{L}$ be a sequence log over elements of $U$ and let $g = g_1 * g_2 * \ldots * g_n$ be a g-sequence, where $g_1, \ldots, g_n \in U$ and $*$ denotes an arbitrary wildcard. For each $i = 1, \ldots, n-1$ and for each $j > i$, the "confidence of $g_j$ towards $g_i$" is the ratio of the number of sequences containing $g_1 * \ldots * g_j$ to the number of sequences containing $g_1 * \ldots * g_i$:

$$confidence_{g_1*\ldots*g_i}(g_j, g_i) = \frac{hits(g_1 * \ldots g_{j-1} * g_j)}{hits(g_1 * \ldots * g_i)}$$

while the confidence of $g_1$ is defined over the whole log as $confidence(g_1, e) = \frac{hits(g_1)}{|\mathcal{L}|}$, where $e$ denotes the empty sequence which is trivially contained in any sequence. □

In this definition, we compute the confidence with which an event $g_j$ occurs after events $g_1, \ldots, g_i$ occur, interleaved with wildcards. For a single event $g_1$, its confidence is equal to its support, in the conventional notion of the term introduced in [ATS93].

It is easy to see that the notion of confidence thus defined subsumes the measure of conversion efficiency. By comparing Def. 8 with Def. 6, we see that:

$$conveff(S, T, \mathcal{G}) = confidence_{\mathcal{G}}(T, S) \qquad (4)$$

where the index denotes an appropriate definition of the group of paths connecting `S` to `T`. Indeed, the miner computes the conversion efficiency as confidence.

**Example 4:** To see how the confidence of the elements of the navigation pattern in Example 3 is computed, we model this pattern as a tree. In particular, we merge all sequences matching $g$ by common prefix. We then annotate each tree node with the number of sequences that contain the tree branch up to that node. The resulting tree is shown in Fig. 2.

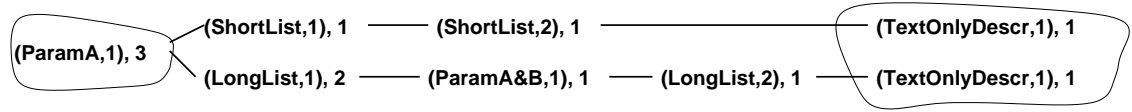

Figure 2: The navigation pattern of (`ParamA,1`) $[0;3]$ (`TextOnlyDescr,1`)

The root of this tree is annotated with the value 3, because (ParamA,1) was invoked in all three sessions. Similarly, both the second and the third session contribute to the annotation of (LongList,1), since both users invoked this format of results after issuing a query with ParamA. This navigation pattern actually consists of two trees, one per page in the g-sequence. The first tree is depicted in Fig. 2. The second one is the degenerate tree, comprising the root node (TextOnlyDescr,1).

For a g-sequence with more than two page occurences, e.g. (ParamA,1)[0;3](TextOnlyDescr,1)[0;2](Descr+Im the navigation pattern would consist of three trees, and the one rooted at ((TextOnlyDescr,1),2) would contain the subsequences leading to (Descr+Image,1).

Thus, the conversion efficiency of (ParamA,1) towards the target page (TextOnlyDescr,1) over the paths matching $g$ is $\frac{1+1}{3} \approx 0.66$. Although there are only two paths matching $g$, the first node of one of the paths has been accessed by three users, one of which did not reach the target page. This must be taken into account when computing the conversion efficiency of ParamA. △

The tree structure depicted in Fig. 2 is an "aggregate tree" [Spi99]. It merges sequences by common prefix and annotates them by the number of appearances of each prefix in the log. This structure is fundamental for WUM: (i) The whole log can be concisely represented as a tree, the permanently stored "Aggregated Log", on which pattern discovery is performed. (ii) A navigation pattern is graphically represented as an aggregate tree, thus permitting the inspection of its contents and the computation of the statistics of nodes within the pattern. The statistics of the non-wildcard elements of the g-sequences are presented separately by the WUM_visualizer service that depicts navigation patterns and their g-sequences graphically [SFW99].

### 4.2 Pattern Templates and Pattern Discovery

A g-sequence and its navigation pattern depict the behaviour of one group of users. In navigation pattern discovery, we seek *all* patterns that have certain properties, such as a minimal frequency within the whole population.

In sequence mining, pattern discovery is usually subject to constraints on minimal frequency and maximal length. In WUM, we allow the specification of more complex constraints on structure, content and statistics of the patterns to be discovered. In particular, the mining language of WUM, MINT, supports the specification of "templates" [SF99]. A template is a vector of *variables* and wildcards, and is accompanied by constraints on the statistics and content of the events (here: page occurences), to which the variables can be bound during mining. Similarly, the wildcards pose structural restrictions on the g-sequences that match the template.

**Example 5:** We assume a fictitious web site, the pages of which have been organized according to the service-based concept hierarchy of Fig. 1. We are interested in the conversion efficiency of pages towards a target page over short paths comprising at most 3 steps between begin and target page. In MINT, we would issue the following query:

```
select t from node as x y, template # x [0;3] y as t
where y.url contains "Descr" and y.occurence = 1
and ( y.support / x.support ) >= 0.2
and x.support >= 30
```

In this query, we specify a template $t$ with two variables $x, y$, thus seeking for g-sequences with two page occurences bound to $x$ and $y$ and at most 3 arbitrary page occurences in between. The symbol # denotes that $x$ should be bound to the first page occurence in a session. This means that we are interested in the conversion efficiency of the entry points in the site.

The variable $y$ should be bound to a target page. According to Fig. 1, we can capture all target pages in one constraint, because they all contain the string "Descr"; for another concept hierarchy, a more complicated regular expression or multiple queries might be necessary. We are further interested in the *first* access to the target page, as denoted by the second constraint on $y$.

The third constraint states that we are only interested in navigation patterns, in which $x$ is bound to a variable showing a conversion efficiency of at least 20% over the short paths. This constraint reflects directly Eq. 4. The last constraint states that $x$ should be bound to site entry pages accessed in at least 30 sessions, in order to avoid patterns that have high confidence because they are very rare. The attribute `support` contains the number of $hits$ for a page occurence (compare with Def. 7). A relative frequency is easily translated to an absolute number.

The result of this query is the set of (g-sequence, navigation pattern) pairs that satisfy all aforementioned constraints. These pairs are built by the miner. △

The analyst uses MINT to specify the templates and constraints that should be satisfied by the patterns to be discovered. The "generalized sequence miner" of WUM, `WUM_gseqm`, takes these specifications as input and performs the discovery, or actually the pattern construction procedure, accordingly. The actual algorithm of `WUM_gseqm` is presented in [Spi99, SFW99]. The software performs best for sessions that are small or have a high degree of overlap. A detailed analysis of the storage requirements, complexity and performance of WUM is provided in [Spi99].

## 5 A Procedure for Evaluating Success on the Basis of the Navigation Patterns

We have thus far described the proposed success measures, the notion of navigation pattern as reflector of page success and the mining mechanism that discovers navigation patterns. This brings us to the last question, namely which patterns should be discovered to compute success values upon.

There is no generic answer to this question. Similarly to other areas of data mining, the knowledge discovery process requires the participation of the human expert, her expertise, background and intuition. In our experiments with real sites, the analysis turned into a highly interactive process, in which each group of patterns discovered by the miner led the analyst to the formulation of the next mining query.

Instead of suggesting a procedure belonging to the sphere of human expertise and intuition, we propose a phase-based procedure to guide the interactive mining process. In particular, we first address the problem of identifying action pages with low contact efficiency. Secondly, we focus on the conversion efficiency for all active sessions, in order to identify pages that cause difficulties for all users. We then perform a comparative analysis between customer sessions and non-customer sessions, in order to identify navigational particularities of each group. These steps are not automated but have to be performed manually.

### 5.1 Evaluating the Contact Efficiency of Action Pages

The *computation* of the contact efficiency and the relative contact efficiency of each action page according to the definitions in 3.2.2 is trivial and can be performed on the basis of the preprocessor data of WUM. MINT queries can also be issued to this end. The *improvement* of the contact efficiency implies the identification of action pages that are rarely reached. This is done with the heuristic of Fig. 3. The example in Fig. 4 illustrates the usage of this heuristic: page C is frequently accessed, but visitors reaching it rarely continue to a target page; this makes page C a candidate for re-design.

---

1. Discovery of all frequent patterns from a site entry page to an action page with low contact efficiency (or relative contact efficiency)
2. Identification of in-between-pages inside these patterns, which are themselves frequent but rarely lead to an action page

   These pages should be redesigned, because they attract many visitors without encouraging them to exploit the site's services.

---

Figure 3: The heuristic method *EVAL_contact_efficiency*

This procedure requires the identification of frequent pages *inside* a pattern. To this purpose, we use the heuristic presented in Fig. 5. This heuristic selects pages that appear frequently *inside* a navigation pattern. Here, the term "frequently" is expressed through a threshold $thr$ that must be given by the analyst.

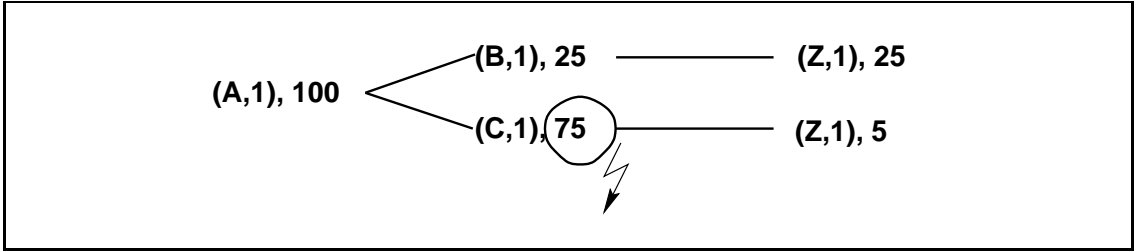

Figure 4: Example for a frequent, but inefficient in-between-page

**Input:** A navigation pattern and a threshold $thr$ for frequence-inside-a-pattern

**Output:** A modified navigation pattern, in which rare subpaths have been removed and their ends have been grouped into nodes appearing more frequently than the threshold $thr$

**Algorithm:** *post_miner*

Traversal of the navigation pattern from the root towards the leaf nodes

For each page $P$ reached during traversal:

- If $P$ is more frequent than $thr$, then retain $P$.
- If $P$ is less frequent than $thr$, then:
    - If there are multiple branches with the same prefix upto $P$,
      then merge them and increase $hits(P)$ accordingly
      else remove $P$

In the first step of this traversal, $P$ will certainly be removed.

Figure 5: An algorithm finding frequent subpatterns within a pattern

The algorithm *post_miner* is currently not supported by WUM and its steps are performed manually. Since it it operates similarly to the miner itself, we intend to implement it as an extension of the miner, so that it can operate as a post miner on the discovered navigation patterns.

## 5.2 Evaluating the Conversion Efficiency for all Active Sessions

The measure of conversion efficiency is applicable to active sessions only. Thus, inactive sessions should be removed at the beginning of this phase, because they can skew the data.

For the evaluation of the conversion efficiency of pages, we distinguish among efficiency (i) over all paths, (ii) over short paths and (iii) over long paths, whereby the notion of "short" and "long" is application-specific and should be determined by the expert in cooperation with the designer. The first two are applicable in sites designed so that each user should be able to reach a target page in a small number of steps. For sites where a long navigation is desirable, all three measures should be considered. In particular, we suggest the heuristic procedure of Fig. 6. The rationale behind this heuristic is that a frequent page may show low conversion efficiency either because it is itself misleading or somehow unintuitive for the users, or because it is frequently followed by some other poorly designed page.

## 5.3 Comparative Analysis of Customer and Non-Customer Sessions

The analysis of the "active log" results in the discovery of pages that may be misleading for all users. However, some patterns may only be frequent among the customer sessions, especially if these sessions constitute only a small portion of the whole active log. Thus, we propose the analysis of the customer sessions only and the comparison of the discovered patterns with those in the non-customer sessions. To this purpose, we partition the active log into the "customer log" comprising the customer sessions and the "non-customer log" containing the non-customer sessions. We then use the heuristic of Fig. 7.

---

1. Removal of the inactive sessions from the log to be tested
   The result is the "active log."

2. Discovery of frequent patterns leading to target pages and computation of the conversion efficiency of their start page
   This step should be performed for all paths and for short paths, and separately for long paths if appropriate for the site.

3. Identification of patterns in which the start page is frequent but the conversion efficiency is low

4. For each such pattern:

   (a) Detection of pages that are themselves frequent but rarely lead to the target page
       The algorithm *post_miner* of Fig. 5 is again used for the identification of pages that are frequent inside the pattern.

   (b) If there are no such pages,
       then the start page must be improved
       else the detected pages must be improved

---

Figure 6: The heuristic method *EVAL_conversion_efficiency*

The heuristic in Fig. 7 first processes the customer log to discover frequent navigation patterns that lead from an action page to a target page. It then compares them with patterns from the non-customer log that start at the same action page and have a similar frequency. The rules according to which two patterns from the different logs are comparable and the mechanism for the comparison of patterns and of paths inside the patterns are presented in [SPF99]. Briefly, a non-customer log pattern is comparable to a pattern from the customer log if their g-sequences have the same prefix, or, more restrictively, are equal in content and length except of the last page occurence. This page occurence would be the target page for a customer log pattern, while in the non-customer log it would be another frequently reached page. Paths in comparable navigation patterns are comparable if they have the same prefix.

The last step of the heuristic in Fig. 7 suggests the inspection of differences in the navigation behaviour between customer and non-customer sessions. Such differences manifest themselves in navigation patterns that appear only in the non-customer log. Also, differences in the relative contact efficiency of the action pages indicate that some action page is preferred in customer sessions more than in non-customer sessions or vice versa.

The comparative analysis of the customer log and the non-customer log can be performed alternatively or subsequently to the analysis of the active log as a whole. The advantage of the comparative analysis lies in the processing of a smaller log, whose sessions reflect a behaviour conformant to the site's goal. This ensures that no patterns of desirable behaviour go unnoticed for not being frequent enough in the entire action log. For the non-customer log, only deviations from this desirable behaviour are of interest. The analysis of the action log as a whole is still necessary, if the customer log is so small that no statistically reliable conclusions can be drawn. A very small customer log also indicates that most of the pages in the site are not successful. An analysis of the entire active log would then reveal the most frequent pages with low conversion efficiency, which should be improved first.

## 6  Improving Success in the SchulWeb Site

We have tested our model in a series of experiments with the "SchulWeb" server (http://www.schule.de). SchulWeb offers access to the most comprehensive database of German schools with their own web sites. It also supports several additional services, including the retrieval of school magazines and access to the *German Educational Resources* (GER) server (http://dbs.schule.de).

With respect to the major service of school search, SchulWeb is organized as an online catalog, thus being similar to a site designed for online merchandizing. Users issue a query by specifying values for one or more search parameters. The query is forwarded to the SchulWeb database and a sequence of pages with

> 1. Processing in the customer log:
>    (a) Evaluation of the relative contact efficiency of the action pages
>    (b) Discovery of frequent navigation patterns from action pages to target pages over all paths, over short paths and, if appropriate, over long paths
>    (c) Selection of the discovered navigation patterns that show high conversion efficiency over short paths and over long paths
>    The conversion efficiency of a customer log pattern over all paths is 1 by definition.
> 2. Processing in the non-customer log:
>    (a) Evaluation of the relative contact efficiency of the action pages
>    (b) For each discovered pattern in the customer log that has a high conversion efficiency:
>       (i) Construction of frequent comparable patterns in the non-customer log
>       (ii) One-to-one comparison of patterns between the two logs to identify where the non-customer pattern deviates from the customer pattern
>       The algorithm *post_miner* of Fig. 5 is used to modify each navigation pattern so that rare pages are removed and paths are merged into more frequent ones.
>    (c) Discovery and inspection of further navigation patterns in the non-customer that
>    start at an action page showing a different relative contact efficiency than in the customer log and
>    are frequent

Figure 7: The heuristic method *EVAL_log_comparison*

lists of schools are generated as a result. These pages also contain the query form to enable users to pose further queries or refine their search. An individual school is selected from a list of schools. SchulWeb supports two formats: the automatically generated description of the school from the SchulWeb database and the homepage of the school.

To improve the success of the SchulWeb, we selected the search for schools as the goal for the analysis. Experimentation with SchulWeb covered all phases of the procedure described in section 5. The study of the conversion efficiency over the whole log is described in [SB00]. An earlier experiment using comparative analysis between the customer log and the non-customer log is reported in [SPF99]. Here, we report on a more recent experiment comparing customer and non-customer patterns, according to the heuristic of Fig. 7. The results of this comparative analysis verified the findings of the analysis of the entire log.

## 6.1 Experimental Settings

The input dataset was a the SchulWeb server log fragment corresponding to one day. The traffic statistics on a weekly basis show that the highest traffic occurs in the days in the middle of the week, so a Tuesday was selected as a typical busy day. The 16 federal states of Germany have different, partially overlapping school holiday periods: The selected day was a holiday in one of the 16 states and a normal school day in the other ones. Subject to the constraint that the log used for the analysis should contain entries of a rather busy working day, the day was selected randomly. After data cleaning, 32,781 page requests were retained in this log.

The data preparation phase encompassed data cleaning, establishment and cleaning of sessions and mapping of the individual URLs to concepts.

*Data cleaning.* Beyond requests for images, we have also removed URL requests originating from non-interactive agents like archivers and software for the automated downloading of pages. To recognize unregistered agents of this kind, we used heuristics: First, we removed clients that always had an empty referrer. Second, we removed clients that accessed consecutive pages in a speed that is too high for human cognition, keeping in mind that SchulWeb is an interactive site.

*Establishment of sessions.* To distinguish among users from the same host or proxy, we exploited the agent log, as suggested in [CMS99]. Further heuristics that exploit the site's topology [CMS99] are not effective for the SchulWeb, because this site corresponds to an almost fully connected graph. The boundaries of a session were defined using an upper limit (of 4 minutes) to the time spent on a single page. Within a session, we have removed consecutive requests for the same URL. Such requests indicate a slow connection and/or an impatient user. They add noise in the patterns without providing any actionable information.

*Mapping URLs to concepts.* We used a service-based hierarchy similar to the generic one shown in Fig. 1 in order to map the dynamically generated URLs of SchulWeb into concepts. At the highest level of the hierarchy, we distinguish among query strategies, which refine our "Action page" concept, school descriptions, which are children of the "Target page" concept, and other pages, which are lists of query results. In SchulWeb, a query invocation is physically the same URL that contains the first page of results. Hence, the concept "Other page" encompasses consecutive pages of query results, as well as URLs related to other SchulWeb services.

An action page is a query strategy. We distinguish among different combinations of search parameters. SchulWeb offers three search parameters: (i) the federal state (acronym: FS) in which the school is located, (ii) the school type (acronym: ST) and (iii) the search for an arbitrary text string (acronym: T), e.g. in the school's name or home town. The first two search parameters can be specified per button click, the last one requires a button click and text input.

A target page is a page describing one school. We distinguish among the school description provided by SchulWeb itself and the homepage of each school. The complete concept hierarchy (in German) is presented in [SB00]. Here, we use translations of all names.

Due to a particularity of the SchulWeb server configuration, the search strategy used in some query invocations could not be identified and was mapped to a dummy strategy. This strategy was consistently ignored in the postmining phase.

*Active sessions and customer sessions.* According to the specification of search strategy invocations as action pages, we identified 1274 sessions of the log as active sessions.

The notion of customer session was based on a finer concept than the simple request for a target page. In particular, an active session was termed a customer session, if a target page was acccessed *and* the stay on the target page exceeded some time threshold (of 7 minutes). We considered this restriction reasonable for a site in which there is a distinction between *inspecting* an object and simply clicking on various objects. We termed such a target page as "/SUCCESS", to distinguish it from target pages also appearing in the non-customer sessions. According to Def. 3, the active log consisted of 725 customer sessions and 549 non-customer sessions.

## 6.2 Analyzing the Customer Log

We have evaluated the conversion efficiency of the search strategies over short paths in the customer log, using the following MINT query:

```
select t from node as a b, template a [0;3] b as t
where a.url contains "SEITE1" and a.occurence = 1
and b.url = "/SUCCESS"
```

Since the invocation of all search strategies in SchulWeb is mapped into a concept of the form `SEITE1-strategy_parameters` it was possible to obtain the navigation patterns of all strategies using a single query. The same query returns the $hits$ of each search strategy, according to Def. 7. From these numbers, the relative contact efficiency of each strategy can also be computed. The efficiency values are shown in Table 1. We only show the results concerning searches for schools in Germany. Searches for schools in other countries were too rare to analyze separately, but their efficiency values were similar.

The values of the relative contact efficiency for the search strategies indicate a preference for regional searches (specification of the federal state), combined with a strong reluctance of the users to invoke strategies that involve the typing of strings. The low relative contact efficiency of the strategy specifying the school type only can be explained through (a) the preference for regional searches and (b) the low selectivity of this strategy, which returns all schools of a specific type from the whole country.

| Search strategy | Acronym | Relative contact efficiency | Conversion efficiency over short paths |
|---|---|---|---|
| Federal state | FS | 35.3 | 75.7 |
| School type | ST | 5.4 | 69.2 |
| **Federal state & School type** | **FS_ST** | **40.5** | **80.6** |
| Text string search | T | 4.7 | 64.7 |
| Federal state & Text string search | FS_T | 2.7 | 78.9 |
| School type & Text string search | ST_T | 7.2 | 67.3 |
| All three parameters | 3P | 7.4 | 85.1 |

Table 1: The efficiency of the SchulWeb search strategies in the customer log

The conversion efficiency of all search strategies over the short paths is high, indicating that customers do reach the schools of interest within very few steps. This was not expected, because strategies not involving text search have low selectivity and return long lists of results.

## 6.3 Comparisons with the Non-Customer Log

For the comparison between customer log patterns and their counterparts in the non-customer log, we have applied step 2 of the mechanism in Fig. 7. We have first computed the relative contact efficiency of the action pages in the non-customer log. The results are shown in Table 2.

| Search strategy | Acronym | Relative contact efficiency |
|---|---|---|
| Federal state | FS | 51.1 |
| School type | ST | 6.1 |
| Federal state & School type | FS_ST | 23.3 |
| Text string search | T | 3.8 |
| Federal state & Text string search | FS_T | 2.0 |
| School type & Text string search | ST_T | 5.6 |
| All three parameters | 3P | 1.8 |

Table 2: The relative contact efficiency of the search strategies in the non-customer log

The relative contact efficiency values again show a preference for regional searches and a reluctance in using search strategies that require text typing. Thus, both customers and non-customers show similar base behaviour. On the other hand, the non-customer log shows a much higher preference for the FS strategy than for the FS_ST strategy, which was the one most frequently used in the customer sessions. The relative contact efficiency of the 3P strategy is smallest among the non-customer sessions, in contrast to customer sessions. We interpret these facts through the high conversion efficiency of FS_ST and 3P: Users that invoke these strategies are more likely to be customers than non-customers.

For the pattern comparison, we have selected the customer log pattern of the FS_ST strategy, because it was the most frequent one and had high conversion efficiency. The comparable patterns were discovered by the following MINT query, in which the German acronym of the FS_ST strategy is used to bind the first template variable:

```
select t from node as x y, template x [0;3] y as t
where x.url endswith "SEITE1-LASALI-D" and x.occurence = 1
and (y.support / x.support) >= 0.045
```

The confidence threshold for this query was purposely low, because the 9 patterns of the result were merged together into a single "pattern" of higher confidence. We removed rare subpaths from this and from the customer log pattern with the *post_miner* algorithm of Fig. 5. The resulting patterns we compared are shown in Fig. 8. In this figure, solid lines indicate original paths, while dashed lines indicate paths produced after merging of subpatterns. We have enumerated the three comparable paths and marked them *subpattern-1* to *-5*. Among the incomparable paths, we identified one of the customer log patterns (*subpattern-4*) and one of the non-customer log patterns (*subpattern-5*). These paths are discussed below.

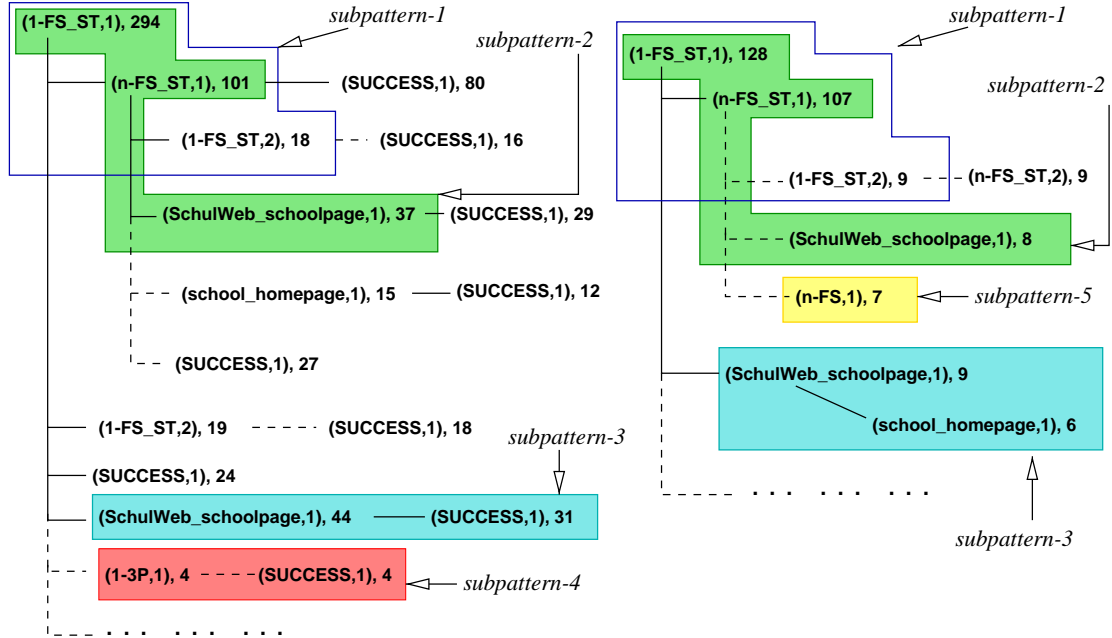

Figure 8: Comparing a customer pattern (left) to the respective non-customer pattern (right)

*Subpattern-1.* In both logs, a number of sessions contain a second invocation of the FS_ST strategy, after two pages of results have been browsed. This occurs in the customer sessions more than twice as frequently as in non-customer sessions (17.8% vs 8.4% of the cases). Instead, non-customer sessions contain a request to a page of results produced with the FS strategy (6.5% of the cases). This path, denoted as *subpattern 5* in Fig. 8, was not followed by customers. Although not immediately apparent from the pattern, this page was produced by an invocation of the FS strategy that *occured before* the pattern we are inspecting.

*Subpattern-2.* In 34.3% of the customer sessions, a second page of results was requested after the first page. In the non-customer sessions, this percentage is as high as 83.5%. Of all customers, 36.6% found a school of interest in this page, reached across a path similar to path (1). Among the non-customer sessions, a school description was requested in only 7.4% of the cases. The rest of the sessions dispersed across different paths.

*Subpattern-3.* After the invocation of the FS_ST strategy, 14.9% of the customer sessions accessed the SchulWeb description of a school. Among the non-customer sessions, this percentage was only 7%, indicating that in non-customer sessions the first page of results did not contain a school of interest. In both logs, the homepage of a school was accessed afterwards, but in non-customer sessions the homepages retrieved were abandoned soon thereafter.

Hence, both logs contain indicators that many users were not satisfied with their search and modified it. While customers insisted on the same strategy, non-customers preferred to browse through some previously retrieved results.

*A customer path using the 3P strategy.* The pattern of the customer log contained a path in which the 3P strategy was invoked. This path was rare, but had the following interesting properties: (i) It always lead to the "/SUCCESS" page. (ii) It invoked the strategy with the highest conversion efficiency in the customer log. (iii) This strategy had the lowest relative contact efficiency in the non-customer log and *never* appeared in the non-customer pattern under consideration.

## 6.4 Interpretating and Exploiting the Results

To interpret our mining results, we combined the observations of the previous subsection with information provided by the site's owner on how SchulWeb is designed and how the search strategies are expected to be used.

*Interpreting the mining results.* Users are reluctant to exploit search strategies that require them to type text. At the same time, many users are interested in schools within a given area. Thus, they select search strategies that permit them to specify at least the federal state of interest without typing, and then they also specify the school type to further restrict the size of the results. However, these strategies have low selectivity: a federal state may have as many as 400 schools. Thus, queries on federal state and school type only tend to produce lists that are by far too long for someone to scroll through.

Hence, whether a user becomes a customer or not depends on whether she finds a school of interest in one of the first pages presenting the results. Moreover, users who browse the results, perform another search with the same strategy or decide to restrict the search by typing text parameters, are more likely to become customers than users who instead step back to results previously retrieved with a less selective search strategy.

The relative contact and conversion efficiencies of the search strategies can be interpreted but were not expected: Under the objective measure of selectivity, the non-customer log is characterized by suboptimal searches while the customer log still contains many searches that are not optimal.

*Acting upon the mining results.* The results of our analysis are actionable. After the interpretation, a concrete modification plan was drawn across the following guidelines.

> Since users prefer regional searches, the conversion efficiency of these searches over short paths should be increased. The selectivity of regional searches will increase if the users specify the name of the town of interest instead of (or next to) the federal state [2].

The search interface of the SchulWeb was modified according to the above plan and a new mining session was launched on a web log sample obtained after the change.

*Modifying the site.* The original query interface of the SchulWeb already provided a large palette of text search options. In particular, the database of schools contains information such as school name, webmaster name, town, names of teachers etc. The visitor can type a text string in a fill-in field *and* specify the attribute, on which the search should be performed. The specification of the attribute is done through a clickable button, which is labelled with the default attribute. If the visitor clicked on the button, all alternative attributes were listed, the town being one among them.

In the original interface, the default attribute was `School name`. After performing the analysis and interpreting the results, the site's owner formed the hypothesis that visitors are not aware that they can search for schools in a specific town, because they do not realize that the clickable button hides alternative search options to the `School name`. Thus, the default attribute, and thus the label of the clickable button was changed into `Town name`, so that the option of searching for schools in a specific town became apparent at first glance.

## 6.5 Studying the Impact of the Site Modification

After modifying the site, we have conducted a further mining session to study the impact of the changes. We have selected the web server log entries of a typical day and established the sessions with the same settings as described in subsection 6.1. We have identified 2414 active sessions, 1272 of them assigned to the customer log and 1142 assigned to the non-customer log.

### 6.5.1 Impact of the GUI Changes on All Active Sessions

To study the impact of the new GUI on the contact efficiency of the search strategies, we have computed the relative contact efficiency (see Def. 5) of each search strategy over the whole sample of active sessions. The results for the original log and for the log after the modification are depicted in Table 3.

---

[2] Some town names are very common, so that one has to specify both the town name *and* the federal state to obtain an unambiguous result.

| Search strategy | Acronym | Relative contact efficiency | | |
|---|---|---|---|---|
| | | *Old log* | *New log* | *Change* |
| Federal state | FS | 42.1 | 32.7 | -9.4 |
| School type | ST | 5.7 | 10.9 | +5.2 |
| Federal state & School type | FS_ST | 33.1 | 30.4 | -2.7 |
| Text string search | T | 4.3 | 7.4 | +3.1 |
| Federal state & Text string search | FS_T | 2.4 | 7.7 | +5.3 |
| School type & Text string search | ST_T | 6.5 | 12.6 | +6.1 |
| All three parameters | 3P | 5.0 | 18.5 | +13.5 |

Table 3: The relative efficiency of the SchulWeb search strategies before and after the modification of the GUI

In Table 3, we see that the relative contact efficiency of all search strategies involving the typing of text has increased dramatically, indicating that the visitors are using the new search option more than they used the old one. This implies that visitors were unaware of the option permitting regional searches in a specific town. Once this option became visible, they were willing to exploit it.

At the same time, the relative contact efficiency of the FS strategy has decreased remarkably. There is also a decrease, albeit less sharp, in the relative contact efficiency of the FS_ST strategy. This is reasonable, since users performing regional searches can search more efficiently by specifying a town name instead of a whole region. The decrease also indicates that regional searches focus on towns rather than big regional areas.

In Table 3, we also observe an increase of the relative contact efficiency of the ST strategy, which retrieves schools by specifying the school type. This indicates a new group of users interested in schools of a specific type and performing superregional searches. The investigation of this new group is beyond the scope of this study.

### 6.5.2 Impact of the GUI Changes on the Customer and the Non-Customer Sessions

The results in Table 3 show that our modification of the SchulWeb search interface has indeed increased the relative contact efficieny of search strategies involving typing of text. Next, we have studied the impact of the modified SchulWeb search interface on the success of the site in turning its visitors to customers. To this purpose, we have computed the relative contact efficiency of each search strategy in the customer and the non-customer log of the new sample, thus partitioning the numbers shown in Table 3 into values for customers and for non-customers. We show the results in Table 4.

| Search strategy | Relative contact efficiency | | | |
|---|---|---|---|---|
| | New customer log | Change | New non-customer log | Change |
| Federal state | 31.5 | -3.8 | 34.1 | -17.0 |
| School type | 10.8 | +5.4 | 10.9 | +4.8 |
| Federal state & School type | 34.0 | -6.5 | 26.4 | +3.1 |
| Text string search | 6.4 | +1.7 | 8.7 | +4.9 |
| Federal state & Text string search | 8.3 | +5.6 | 7.0 | +5.0 |
| School type & Text string search | 13.1 | +5.9 | 12.0 | +6.4 |
| All three parameters | 17.9 | +10.5 | 19.2 | +17.4 |

Table 4: The relative efficiency of the SchulWeb search strategies in the customer log and the non-customer log of the modified site

In the third row of Table 4, we see that the number of customers using the FS_ST search strategy decreased, while the respective number of non-customers increased. This means that visitors opting for this strategy are less likely to perform successful searches (and therefore to appear in the customer log). As already noted, the use of a text search option would be a more appropriate alternative. However, the values in the Table show that the increase in the relative contact efficiency for the text-based search strategies is rather higher in the non-customer log than in the customer log. This indicates that visitors are not very successful in exploiting the text-based strategies they select. To understand this indication better, we have also computed the conversion efficiency of the search strategies over short paths in the customer log, using the same mining query settings as described in subsection 6.2. The results are depicted in Table 5.

| Search strategy | Acronym | Conversion efficiency over short paths | |
|---|---|---|---|
| | | New customer log | Change |
| Federal state | FS | 79.6 | +3.9 |
| School type | ST | 76.8 | +7.6 |
| Federal state & School type | FS_ST | 81.9 | +1.3 |
| Text string search | T | 85.2 | +20.5 |
| Federal state & Text string search | FS_T | 80.0 | +1.1 |
| School type & Text string search | ST_T | 80.7 | +13.4 |
| All three parameters | 3P | 78.9 | -6.1 |

Table 5: The conversion efficiency of the SchulWeb search strategies in the customer log of the modified site

The last column of Table 5 shows that the conversion efficiency over short paths has increased for all strategies except of the one involving all three parameters (3P). The percental improvement of the ST strategy should be attributed to the strategies refining it, namely FS_ST and ST _T: A query for all schools of a specific type has a very low selectivity; refining it by specifying the federal state or the desired town of location increases the selectivity factor considerably.

The percental improvement for the FS, FS_ST and FS_T strategies in Table 5 is marginal. It can be the result of noise. Another plausible explanation is that it relates to the usage of the 3P strategy. In particular, the contact efficiency of the 3P strategy has increased, but its conversion efficiency has decreased, indicating a larger number of failed searches. Some of these searches were continued by changing the search strategy to one of FS, FS_ST and FS_T. A percentage of them lead to success, contributing to the improvement of the conversion efficiency of these strategies.

It can be attributed to the increasing experience of the SchulWeb users and to the failed searches caused by the invocation of the 3P strategy: Visitors specifying all search parameters and obtaining no results are likely to specify less parameters in their new query.

Finally, the percental improvement of the conversion efficiency for the strategies T and ST_T is quite high. This fact, when combined with the marginal improvement of the FS_T strategy, indicates that visitors specify the desired town instead of the whole region. The high conversion efficiency of these strategies cannot be attributed to a strategy refinement, because the conversion efficiency of the 3P strategy has decreased, nor to a strategy change, because the improvement of the other strategies is only marginal. Hence, search strategies involving the specification of a town name are quite successful, thus justifying the modification of the site to support this kind of search explicitly.

On the other hand, visitors combining all three search parameters are less successful in their searches. A possible explanation is that queries with the 3P strategy are issued by users who intend to specify the region, the school type and the school name (instead of the town) and are confused by the modified interface.

### 6.5.3 Resumée of the GUI Changes

The comparative experiments on the navigation patterns before and after the modification of the query engine interface show that the findings of WUM have lead to measurable improvements of the site. In particular, the first analysis of the customer and the non-customer behaviour has shown that users are mainly interested in regional searches but they perform them suboptimally in the sense that they obtain too long result lists. Long lists confused some users; while some of them were willing to browse through several pages of results, others gave up. The analysis has further revealed the origins of suboptimality, namely the fact that selective search strategies were rarely used. Finally, the discovered navigation patterns indicated how the problem could be resolved, i.e. by increasing the awareness of the users on the existence of highly selective search strateties for regional searches.

The user interface to the query engine of SchulWeb was modified according to these findings. The analysis of the navigation patterns of all users and of customers versus non-customers after the change has shown that users became aware of the existence of search strategies that return optimal results and started to use them.

# 7  Conclusions and Outlook

In this study, we have presented a model improving the success of a web site with data mining techniques. When designing this model, we have posed the following requirements on its functionality: (i) It should be able to take all users of the site into account, not only a selected sample. (ii) It should be appropriate for continuous testing on the site, i.e. it could not rely on off-line experiments or direct interaction with the users. And most importantly, (iii) it should measure the site's success *and* provide indications on how the success can be maximized.

Our model satisfies all these requirements. We measure the success of a site as the efficiency of its pages in motivating the users to exploit the supported services and acquire the offered goods. We have defined three measures in this context, the contact efficiency, relative contact efficiency and conversion efficiency of a *page*. This enables the analyst to identify the impact of each page on the success of the site and decide which pages should be improved. Our measures are based on concepts used in marketing, but are defined in such a generic way that they are applicable on a large variety of sites.

To evaluate the efficiency values of a site's pages, we analyze the navigational behaviour of the site's visitors with the web usage miner WUM. WUM provides a powerful notion of navigation pattern and an expressive mining language, with which the efficiency measures can be expressed, and the navigation patterns underlying them can be inspected.

Deciding how the pages of a site should be improved requires an understanding of the users' navigation patterns. We propose a heuristic method, according to which the analyst can proceed in preparing the site for web usage mining, discover appropriate navigation patterns in the mining phase, evaluate the efficiency values, inspect these patterns in a postmining phase and draw conclusions on which pages need which improvements. Our heuristic offers two options: The analyst may decide to perform navigation pattern discovery over the entire log or to split it into a customer log and a non-customer log and perform a comparative analysis of the two. Thus, differences between desired and non-desired behaviour can be identified and dealt with.

We have tested our model in a real web site. Although not of commercial nature, this site is organized as an online catalog of products and can thus be observed as a representative of sites designed for online merchandising. Our experiments led to a better understanding of the interests and behaviour of the site's visitors and to a concrete plan of how the site should be improved to better fit its visitors' interests and expectations. The modification of the site led to an improvement of the contact efficiency of the search strategies supported by the site and to an increased conversion efficiency for most of them.

Extensions of our model proceed in several directions. First, we will support the construction of concept hierarchies by a convenient user interface. Currently, this task is done by application specific scripts. Second, we will add a batch mode with a query generator for more efficient over-night processing. Third, we are working on additional measures that reflect further characteristics of site usage. Fourth, we are interested in combining knowledge about user profiles with our navigation patterns. Further, we are extending the miner WUM to better support the postmining phase, which is actually performed manually. Finally, we plan to refine the heuristic procedure we propose for improving the success and to provide as many generic guidelines for this complex process as possible.